\pdfoutput=1

\documentclass[11pt]{article}

\usepackage[final]{mtsummit25}
\usepackage{copyright}

\usepackage{times}
\usepackage{latexsym}

\usepackage[T1]{fontenc}

\usepackage[utf8]{inputenc}

\usepackage{microtype}

\usepackage{inconsolata}

\usepackage{graphicx}
\usepackage{tikz}
\usetikzlibrary{positioning, shapes.geometric, arrows}

\usepackage{romannum}
\usepackage{amsmath}
\usepackage{multirow}
\usepackage{subcaption}
\usepackage{pgfmath}
\usepackage{CJKutf8}
\usepackage[]{todonotes}
\usepackage{xcolor}

\title{Gender-Neutral Machine Translation Strategies in Practice}

\author{Hillary Dawkins \And Isar Nejadgholi  \\
  Digital Technologies Research Centre\\National Research Council Canada (NRC-CNRC) \\
  \texttt{\{hillary.dawkins, isar.nejadgholi, chikiu.lo\}@nrc-cnrc.gc.ca} \\
  \And Chi-kiu Lo \begin{CJK*}{UTF8}{bsmi}羅致翹\end{CJK*}}

\begin{document}
\maketitle
\begin{abstract}
Gender-inclusive machine translation (MT) should preserve gender ambiguity in the source to avoid misgendering and representational harms. 
While gender ambiguity often occurs naturally in notional gender languages such as English, maintaining that gender neutrality in grammatical gender languages is a challenge. 
Here we assess the sensitivity of 21 MT systems to the need for gender neutrality in response to gender ambiguity in three translation directions of varying difficulty.  
The specific gender-neutral strategies that are observed in practice are categorized and discussed. Additionally, we examine the effect of binary gender stereotypes on the use of gender-neutral translation.
In general, we report a disappointing absence of gender-neutral translations in response to gender ambiguity. 
However, we observe a small handful of MT systems that switch to gender neutral translation using specific strategies, depending on the target language.  
\end{abstract}

\section{Introduction}


Gender bias in language technologies has long been a concern for the community (see \citet{sun-etal-2019-mitigating} for an overview). However, machine translation (MT) stands out as a particularly important area of study due to its widespread use, its potential to foster global connections and uplift underserved communities, coupled with the unique technical challenges posed by grammatical gender differences across languages.
Research on gender bias in machine translation often examines misgendering errors resulting from stereotypes or the default use of masculine forms \cite{savoldi-etal-2021-gender}. These errors contribute to both representational harms -- such as the underrepresentation of female and non-binary individuals, or the limited depiction of women in certain roles \cite[][\textit{inter alia}]{barclay-2024, stanovsky-etal-2019-evaluating, troles-schmid-2021-extending, solmundsdottir-etal-2022-mean} -- and allocative harms, for example the increased burden on female users to manually correct machine-translated text \cite{savoldi-etal-2024-harm}. Based on a human user study, \citet{dev-etal-2021-harms} found that such misgendering errors are particularly harmful for non-binary individuals. Here we are concerned with misgendering harms that arise, not due to stereotypes, but rather due to ambiguous or underspecified gender in the source text, a situation that occurs frequently in English. 

One obvious solution is to use gender-inclusive language (i.e., language that encompasses all gender identities when the gender of a particular referent is unknown, or when gender is unimportant). However, the ease of this strategy varies by language.
In notional gender languages, such as English, gender has minimal effects, primarily manifesting through gendered pronouns (e.g., he, himself) and relatively few gender-specific nouns (e.g., brother/sister, husband/wife, waiter/waitress, etc.). 
English also features some androcentric language (e.g., mankind) and, depending on the text domain and social norms, lends itself to the use of masculine generics when gender is irrelevant (e.g., ``any driver in violation will lose his license'').
It is easy to see then why gender inclusive language in English is relatively straightforward; 
the widely established use of ``they/themselves'' as a singular personal pronoun and the existence of in-vocabulary gender-neutral nouns (e.g., sister/brother $\rightarrow$ sibling, husband/wife $\rightarrow$ spouse, waiter/waitress $\rightarrow$ server, mankind $\rightarrow$ humankind) enable neutrality with minimal semantic or lexical perturbations. 
However, in grammatical gender languages (Spanish, Italian, French, Icelandic, Czech, etc.), multiple parts of speech typically need to agree with the gender of the referent (including nouns, verbs, and adjectives), and furthermore, grammatical gender may be restricted to binary (masculine/feminine) cases. 
Therefore, gender neutrality can require more contrived rewrites, potentially using new or out-of-vocabulary words, and coherence may be sacrificed \cite{piergentili-etal-2023-theory}.  
In the context of machine translation, a task \textit{designed} to preserve meaning and coherence, a difficulty naturally arises in translation from lower to higher grammatical gender agreement contexts.

Gender neutral translation is the task of preserving gender neutrality in response to gender ambiguity in the source. 
Here, we study machine translation from English into grammatical gender languages, and assess MT systems' \textit{sensitivity} to the need for gender neutrality. 
We ask, if appropriate, low-barrier gender-neutral translation options exist in the target language, will modern MT systems use those options \textit{in response} to gender ambiguity? We find that a deliberate form of gender ambiguity in English (using ``they'' as a singular personal pronoun) triggers a gender-neutral response in a small handful of surveyed MT systems, but overall, we observe a disappointing insensitivity to gender ambiguity. The results indicate much room for improvement, and we release the scoring code for the test set to enable future work at \href{https://github.com/hillary-dawkins/wmt24-gender-dialogue}{https://github.com/hillary-dawkins/wmt24-gender-dialogue}.

\section{The Task of Gender Neutral Translation}

Based on the assertion that gender ambiguity should be preserved through translation to mitigate harm, 
\textbf{gender-neutral translation} (GNT) is an emergent challenge task to benchmark MT systems for this behaviour \cite{piergentili-etal-2023-theory}. 
Through a careful analysis of existing gender-inclusive language guidelines for both English and Italian, and a survey of human participants, \citet{piergentili-etal-2023-theory} define the desiderata of gender neutral translation: Neutrality should be maintained in the target if gender is ambiguous given the source, but conversely, gendered language should be used if it is known given the source. This makes gender neutrality a dynamic constraint, meaning MT systems should be \textit{sensitive} to the need of it. 
In this study, we define a test set and metric to measure this sensitivity, termed the \textbf{gender-neutral response}, in a controlled way. 
Our test instances are templated to create pairs of inputs that differ only in whether the referent's gender is known or unknown, and adjective translations in target gendered languages are used to measure the sensitivity.  

Prior studies in gender-neutral translation have been somewhat limited in scope; there has been a focus on translating gender-neutral pronouns \cite{cho-etal-2019-measuring, barclay-2024}, including neo-pronouns \cite{lauscher-etal-2023-em}, and gender-neutral nouns that are prone to masculine generic translation \cite{savoldi-etal-2023-ines}.
These studies are focused on a first-order effect (i.e., the gender of the referent word itself -- a pronoun or noun), whereas we are interested in second-order gender effects -- the gender agreement of an adjective with its resolved referent.
This is representative of the challenge that occurs when translating from English to grammatical gender languages.   

\citet{piergentili-etal-2023-hi-GeNTE} was the first, to our knowledge, to create a more robust, natural GNT benchmark (GeNTE), involving real in-the-wild inputs with reference translations (English $\rightarrow$ Italian). The test set contains both gender-ambiguous and gender-determined inputs paired with both gendered and gender-neutral reference translations.   
Using the GeNTE benchmark, \citet{savoldi-etal-2024-prompt} found that both neural MT systems and large language models (LLMs) struggle with gender-neutral translation, but we suggest this may be due to the difficulty of the test set: multiple gender agreements are needed -- or should be avoided -- in each output, and the target language, Italian, has only binary gender grammatical cases in its formal form. Furthermore, because the inputs are derived from real, naturally occurring language, it is hard to control for the existence (or difficulty) of a coherent gender-neutral translation.  

Here, we take a step back and provide a somewhat easier and more controlled test set for GNT to compliment the GeNTE. Uniquely, the construction of the paired inputs, differing only in determined/ambiguous gender, allows us to measure a sensitivity to the need for gender neutrality -- a prerequisite for GNT. 
In our setup, all gendered translations are measured via adjectives, and this provides several advantages. Firstly, the choice of gender declension after translation can be determined via dictionary lookup, rather than a learned classifier as in \citet{piergentili-etal-2023-hi-GeNTE} and \citet{savoldi-etal-2024-prompt}. The dictionary method allows for a more fine-grained analysis of specifically which gender-neutral strategies are being used in translation (e.g., a gender-neutral adjective vs.\ a noun phrase). Secondly, adjectives in our target languages typically have gender-neutral synonyms available, and therefore the barrier to obtaining a gender-neutral translation is low (i.e., a human translator is not expected to struggle with this task). We include three target languages with different grammatical gender cases applied to adjectives: 
\begin{enumerate}
    \item Icelandic and Czech: Have a grammatical gender neuter case that does not usually apply to people.\footnote{Though dynamic social norms may permit this to varying degrees.}
    \item Spanish: Has only masculine and feminine grammatical gender cases (in the standard/formal form of the language). 
\end{enumerate}
Although using templates to study gendered adjectives in isolation does not represent the breadth of real-world complexities, it does provide a starting point grounded in gender resolution that is typical of grammatical gendered languages.

\section{Test Suite and Participating MT Systems}


We make use of a test suite released as part of a shared challenge task at WMT24\footnote{https://www2.statmt.org/wmt24/index.html}: Gender Resolution in Speaker-Listener Dialogue Roles \cite{dawkins-etal-2024-wmt24}. All instances in the test suite involve spoken dialogue and meta-context surrounding the dialogue. Most instances involve two characters, and both characters may take on the speaker or listener role at times. Adjectives within spoken dialogue either refer to the speaker (self-referential) or the listener, and
meta-context either resolves the gender of the adjectives' referents, or leaves the gender ambiguous. As a minimal example: \texttt{$\langle$``I'm/you're stubborn'', I said to him.$\rangle$} either resolves the gender (\texttt{you're}) or not (\texttt{I'm}).
Examples of test suite instances supporting the current study are shown in Figure \ref{fig:templates}, and a full description of the test suite is provided in the Appendix. 

The WMT24 shared task \cite{kocmi-etal-2024-findings} garnered 40 unique MT systems operating in 11 translation directions, and the gender resolution test suite obtained translations from 21 unique MT systems in 3 language directions (EN $\rightarrow$ IS, CS, ES). 
Participating MT systems include commercial online systems, dedicated neural MT systems, and large language models (LLMs) prompted for MT. Throughout, results use the system names as provided by the shared task organizers (i.e., commercial online systems are anonymized). 
One key challenge in gender-neutral translation, as highlighted by \citet{piergentili-etal-2023-theory}, is the need for paragraph-level inference to accurately determine gender. This consideration makes the WMT24 submissions particularly relevant and timely to study. Both the general MT shared task (formerly known as the News Translation task) and several challenge tasks placed an emphasis on paragraph-level translation. Among the 28 participating systems that specified their translation strategies, 17 reported using a paragraph-level or hybrid approach.

\begin{figure*}[t]
\centering
  \includegraphics[scale=.9]{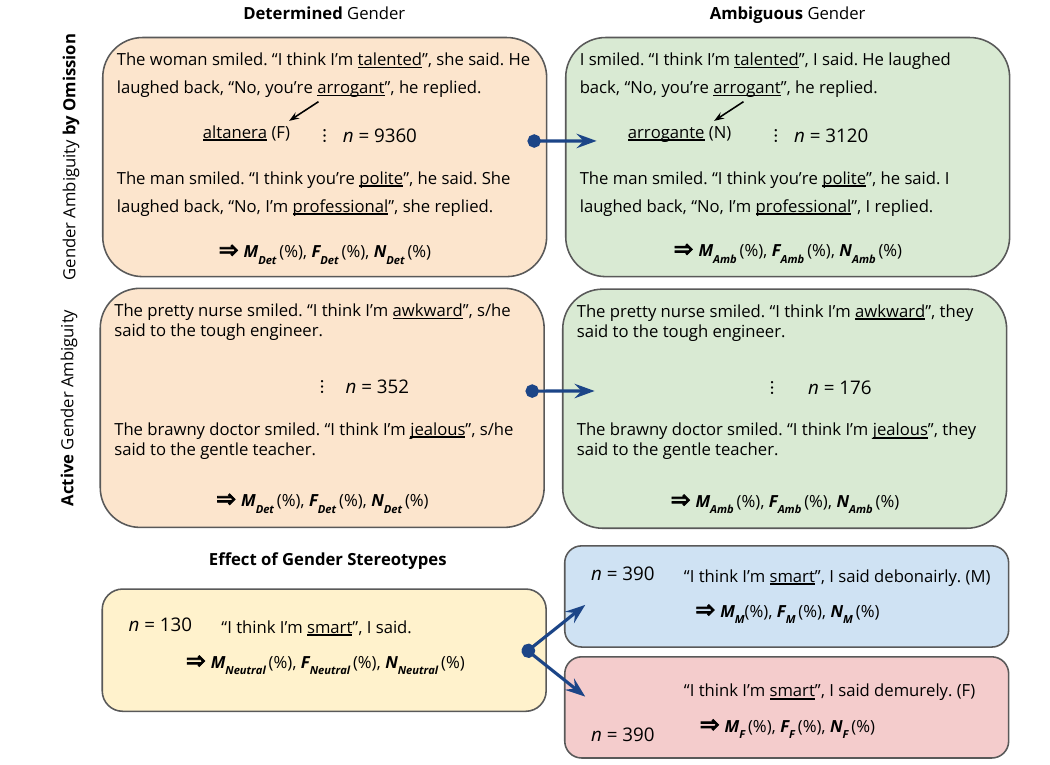}
  \caption{Examples of source inputs within the test suite. All inputs use adjectives to refer to either the speaker or the listener of known or unknown gender, given the context. Each determined gender template (orange) has a matching pair input with a small perturbation applied to make the gender ambiguous (green). The default masculine response $\Delta M = M_{Amb} - M_{Det}$ and the gender-neutral response $\Delta N = N_{Amb} - N_{Det}$ measure the effect of gender ambiguity on the resulting translation. Gender stereotyped adverbs (blue, red) are sometimes used to influence the assumed gender of the speaker (yellow). $n$ represents the number of data points in each category.}
  \label{fig:templates}
\end{figure*}

\section{Observed Gender-Neutral Strategies}
\label{sec:N_types}

Over the entire test suite of 17,966 source adjectives, each translated into three target languages involving 21 unique MT systems, we observed 5 categories of gender-neutral translation strategies in practice: 

\begin{enumerate}
    \item \textbf{Gender-neutral adjectives} $(N_1)$: The translated adjective takes the same inflection in either the feminine or masculine grammatical case. 
    \item \textbf{Neuter case adjectives} $(N_2)$: For adjectives that do not have the same form for masculine and feminine cases, the grammatical gender neuter case is used if it exists (as in Icelandic and Czech). 
    \item \textbf{Other gender-neutral parts of speech} $(N_3)$: Adjectives in the source are translated as a noun phrase or another part of speech without the need for gender agreement in the target language.  
    \item \textbf{English substitution} $(N_4)$: The source adjective is directly copied in translation.  
    \item \textbf{New or alternative morphology} $(N_5)$: New morphology is used to represent multiple possible variants of a gendered adjective in the target language. The observed variations mostly accommodate binary grammatical gender only\footnote{Technically, alternative morphology with only binary options is not a fully gender-neutral strategy, but it is included here to measure an attempt at gender-neutral options in response to gender ambiguity. See Section \ref{sec:discussion} for further discussion.}.  
\end{enumerate}

Table \ref{tab:N_types} provides an example of each gender-neutral translation strategy for each target language.

\begin{table*}
  \centering
  \begin{tabular}{lllllll}
    \hline
    \textbf{Source} & \textbf{Target} & \multicolumn{5}{c}{\textbf{Gender-Neutral Strategies}} \\
    \textbf{Adjective} & \textbf{Language} & $N_1$ & $N_2$ & $N_3$ & $N_4$ & $N_5$  \\
    \hline
    fit & Spanish & fuerte  & --- & en forma  & fit & musculos(o/a) \\
    nonsensical & Czech & absurdní & nesmyslné & nemám smysl  & nonsensical & nesmysln(ý/á) \\
    cautious & Icelandic & varkár & varkárt & á varðbergi & cautious & 	huglítil(l) \\ 
    \hline
  \end{tabular}
  \caption{Examples of each gender-neutral translation type by target language.}
  \label{tab:N_types}
\end{table*}


\section{Definitions and Metrics}

We define the \textit{gender neutrality} of an MT system on a specified subset of source adjective inputs as the proportion of gender-neutral translations observed in the output. 
\textit{Gender-determined} inputs refer to source sentences that provide unambiguous gender information, such that the gender of the adjective referent is known. In contrast, \textit{gender-ambiguous} inputs do not contain enough information to resolve the gender. For English source inputs, gender ambiguity naturally arises due to gender-neutral first- and second-person pronouns (I, you), and can also be induced through the use of ``they'' as a singular third-person pronoun. We refer to the former as gender ambiguity \textit{by omission}, and the latter as \textit{active} gender ambiguity since it involves a more deliberate choice to avoid ``he/she'' (and especially to avoid a default masculine form if gender is unknown or unimportant).    

Here we are interested in how MT systems \textit{respond} to a need to preserve gender ambiguity from source to target, and therefore we define metrics to measure the change in gender neutrality when source inputs switch from determined to ambiguous gender. Firstly, \textbf{baseline gender neutrality} $(N_{Det})$ measures the proportion of gender neutral translations in gender-determined cases, and provides a sense of how various MT systems make use of the gender-neutral strategies by default. 
Baseline gender neutrality is naturally expected to vary by target language due to variable availability, coherence, and ease of the gender-neutral strategies. 
When gender is made ambiguous, all else being equal, the \textbf{default masculine response} measures the increase in the proportion of masculine-form translations $(\Delta M = M_{Amb} - M_{Det})$, and the \textbf{gender-neutral response} measures the increase in the proportion of gender-neutral translations $(\Delta N = N_{Amb} - N_{Det})$. Refer to the examples in Figure \ref{fig:templates} illustrating the move from determined to ambiguous gender in the source sentences.

Note that, depending on the context, either high baseline gender neutrality or high gender-neutral response (or some combination) could be defined as the ideal translation behaviour.\footnote{Although, based on the desiderata defined by \citet{piergentili-etal-2023-theory}, only high gender-neutral response would be ideal.}

\section{Observations}

\subsection{Frequency of gender-neutral translations}

\textbf{Baseline gender neutrality} is dominated by gender-neutral adjectives across languages and MT systems (refer to Appendix Table \ref{tab:N_breakdown_all}). As expected, the frequency varies by language due to a natural disparity in the availability of gender-neutral adjectives (on average, 39\% of translations are gender neutral in Spanish, compared to 19\% for both Czech and Icelandic)\footnote{Only 350 adjectives are included in the test suite -- therefore these proportions are not likely representative of the availability of gender-neutral adjectives in the three target languages as a whole. This comparison only serves to illustrate that different baseline gender neutrality levels among target languages are not unexpected.}. However, variance across MT systems is high within languages, suggesting that some systems strongly prefer the use of gender-neutral translations (e.g., in Icelandic, baseline gender neutrality ranges from 13\% to 36\%).

\subsection{Strong masculine default response}

When source sentences change from determined gender to ambiguous gender (Figure \ref{fig:templates} orange to green), the \textbf{default masculine response} measures the increase in output masculine translations. 
The majority of surveyed MT systems have a strong tendency to default to a masculine translation under gender ambiguity in the source. 
This response occurs both when the gender is ambiguous by omission (refer to Table \ref{tab:I_all} $\Delta M$), and when the gender is actively ambiguous (Table \ref{tab:they_all} $\Delta M$).

\textit{Any} response in the output translation when moving from determined to ambiguous gender implies a sensitivity to the need for gender agreement, although, arguably, defaulting to a masculine form is not the best use of that awareness. The consequence of a default masculine response is fewer feminine translations under ambiguous gender, as shown by the change $\Delta F$ in Tables \ref{tab:I_all} and \ref{tab:they_all}.  

\subsection{Gender-neutral sensitivity}

We now arrive at the main result of the paper, which is to determine whether modern systems respond to the need for gender ambiguity in translation by employing the gender-neutral strategies. The so-called \textbf{gender-neutral response} measures the increase in the use of such strategies when source sentences are switched from determined to ambiguous gender. 

Unfortunately, we observe \textit{no significant gender-neutral response across any MT system or language} when the gender is made ambiguous by omission (I, you). All absolute changes in the proportion of gender-neutral translations are within $\pm3\%$ (refer to Table \ref{tab:I_all} $\Delta N$).
This implies that although some systems prefer the use of gender-neutral translations, that preference is \textit{not in response} to a need for gender neutrality. 

\begin{table*}
  \centering
  \begin{tabular}{llrrrrrrr}
    \hline
    \textbf{Lang.} & \textbf{System}& $(M, F, N)_{Det}$ & $\Delta N$ &$\Delta N_1$ & $\Delta N_2$ & $\Delta N_3$ & $\Delta N_4$ & $\Delta N_5$  \\
    \hline
    \multirow{ 2}{*}{IS} & Claude-3.5 & (0.42, 0.36, 0.22) & \textbf{0.180} & -0.015 & -0.012 & -0.009 & 0.000 & \underline{0.215} \\
     & Aya23 & (0.50, 0.20, 0.30) & \textbf{0.078} & -0.003 & \underline{0.047} & 0.039 & -0.006 & 0.000 \\
     \hline
    \multirow{ 3}{*}{CS} & CUNI-GA & (0.40, 0.40, 0.20) & \textbf{0.081} & \underline{0.056} & 0.000 & 0.017 & 0.009 & 0.000 \\
     & ONLINE-W & (0.39, 0.41, 0.19) & \textbf{0.073} & \underline{0.070} & 0.000 & -0.005 & 0.007 & 0.000 \\
     & Unbabel-Tower & (0.40, 0.40, 0.19) & \textbf{0.073} &\underline{0.050} & -0.003 & 0.022 & 0.003 & 0.000 \\
    \hline
  \end{tabular}
  \caption{The 5 machine translation systems with a \textbf{non-zero gender-neutral response} observed in specific target languages -- out of 21 systems in 3 language directions. The specific type of gender-neutral strategy contributing the most to the response is underlined. For example, when gender is known given the source, Claude-3.5 produces gender-neutral translations in Icelandic at a baseline rate of 22\% ($N_{Det}$). When gender is made ambiguous in the source, this increases to 40\% ($\Delta N = 0.180$), primarily through the use of alternative morphology ($N_5$).}
  \label{tab:they_response}
\end{table*}

However, we observe that more active gender ambiguity in the source (they) does trigger a gender-neutral response for a small handful of MT systems in Czech and Icelandic (refer to Table \ref{tab:they_response}).
Both within languages and across languages, the specific gender-neutral strategies driving the response differ. 
In Icelandic, the MT system with the strongest response (Claude-3.5) reacts to gender ambiguity by using alternative morphology ($\Delta N_5 = +22\%$), whereas Aya23 reacts by using the gender neuter case and other gender-neutral parts of speech. 
In Czech, all three systems with a significant response switch to using gender-neutral adjectives, a response that is never invoked in either Icelandic or Spanish. 
Furthermore, the response is inconsistent for specific systems across languages. 
For some gender-neutral strategies, this may be explained by divergent social norms encoded in the target language (e.g., Aya23 uses the neuter case in Icelandic, but not in Czech), but not so for other strategies. 
For example, Claude-3.5 uses alternative morphology frequently in Icelandic (e.g., ``(ur)'' appending regular adjectives to denote both masculine and feminine cases), but seldom in the higher resource languages Czech or Spanish, despite an observed capacity to do so (e.g., ``o/a'' and ``ý/á'' are both observed infrequently). 
The full results for all systems and languages are shown in the Appendix Table \ref{tab:they_all}.

\subsection{Gender neutrality and stereotypes}
\label{sec:stereotypes}

\begin{table*}
  \centering
  \begin{tabular}{lrrrrr}
    \hline
     \textbf{System (Lang.)}& $(M, F, N)_{Neutral}$ & $(M, F, N)_{StereoM}$ & $(M, F, N)_{StereoF}$ & $\Delta G_{avg}$ & $\Delta N_{avg}$\\
    \hline
    GPT-4 (IS) & (0.29, 0.48, 0.23) & (0.47, 0.32, 0.21) & (0.24, 0.54, 0.22) & \textbf{0.119} & -0.014 \\                
    Claude-3.5 (CS) & (0.64, 0.18, 0.17) & (0.78, 0.06, 0.16) & (0.48, 0.35, 0.17) & \textbf{0.147} & -0.005 \\
    ONLINE-W (ES) & (0.48, 0.16, 0.36) & (0.51, 0.12, 0.36) & (0.28, 0.36, 0.36) & \textbf{0.116} & 0.001 \\
    \hline
  \end{tabular}
  \caption{
  Effect of binary \textbf{gender stereotypes} on the proportion of masculine $(M)$, feminine $(F)$, and gender-neutral translations $(N)$. Subscripts denote the stereotype influencing the assumed gender of the adjective referent (either none, masculine, or feminine). 
  The average effect on binary gender is denoted by $\Delta G_{avg}$, and the average effect on the proportion of neutral translations is denoted by $\Delta N_{avg}$.
  Here, the binary gender proportions are significantly impacted by stereotype cues (higher $\Delta G_{avg}$), while the proportion of gender-neutral translations is roughly constant ($\Delta N_{avg} \approx 0$). 
  The full results are shown in the Appendix Table \ref{tab:stereo_all}.
  }
  \label{tab:stereo_subset}
\end{table*}

\begin{figure}[t]
\centering
  \includegraphics[scale=.62]{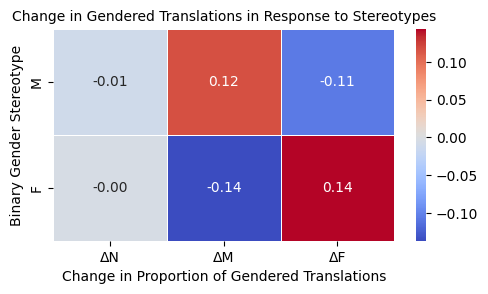}
  \caption{The average trade-off in the proportions of gendered translations (neutral, masculine, feminine) in response to the stereotyped adverbs (male or female), for the most affected MT systems (Table \ref{tab:stereo_subset}). $\Delta$ values represent absolute changes compared to the baseline levels with no stereotype present. Binary gender stereotypes affect the proportions of masculine and feminine translations, and have no effect on the neutral translations (Section \ref{sec:stereotypes}).}
  \label{fig:heatmap_stereo}
\end{figure}

Lastly, we observe that gender neutrality is not affected by binary gender stereotypes. That is, for each MT system and target language, the frequency of gender-neutral translations is consistent across source inputs (\romannum{1}) without any gender stereotypes (e.g., Figure \ref{fig:templates} yellow box), and (\romannum{2}) with binary gender stereotypes influencing the assumed gender of an adjective referent (e.g., Figure \ref{fig:templates} blue and red boxes).   

Gender stereotypes are known to have a significant impact on the gender resolution of adjective referents, as measured by masculine vs.\ feminine adjective agreements. For example, ``The nurse is \underline{talented}'' $\rightarrow$ ``talentos\underline{a}'' (F) vs.\  ``The doctor is \underline{talented}'' $\rightarrow$ ``talentos\underline{o}'' (M).
Though the trade-off between masculine and feminine cases in response to stereotypes is well known (see \citet{dawkins-etal-2024-wmt24} for a fuller analysis of stereotype effects within this test suite), here we include the proportion of neutral translations in the analysis. The results indicate that the trade-off occurs between the binary gender translations only, while the presence of the stereotype does not perturb gender-neutral translations (refer to the results in Table \ref{tab:stereo_subset}). Figure \ref{fig:heatmap_stereo} illustrates the average trade-off between the gendered translations for the most affected systems. 
This implies that the base adjective choice is the first-order translation decision, and the gender agreement is second-order. 
This is consistent with the observed difficulty MT systems have in switching to a gender-neutral adjective in response to a need for gender neutrality; only three MT systems successfully use this strategy, and furthermore, it is only invoked under active gender ambiguity, as discussed in the previous section.  

\section{Discussion}
\label{sec:discussion}

\begin{figure*}
\setlength{\dimen1}{\dimexpr 0.5\textwidth-0.5\columnsep}
\sbox0{\includegraphics[width=\dimen1]{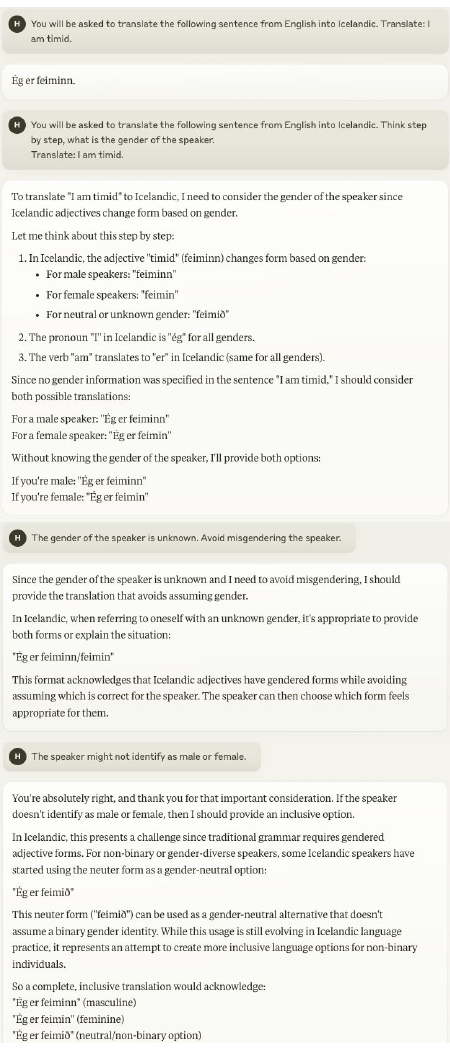}}
\sbox1{\includegraphics[width=\dimen1]{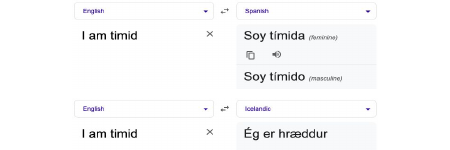}}%
\sbox2{\includegraphics[width=\dimen1]{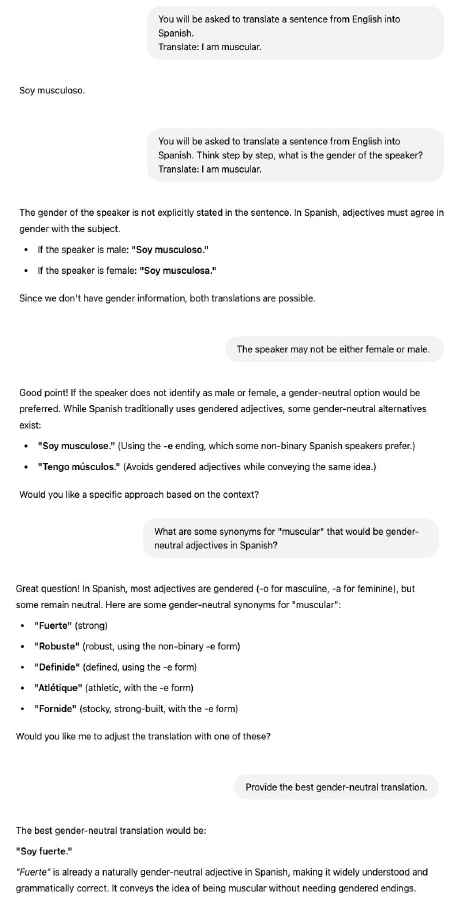}}%
\sbox3{\begin{subfigure}[b]{\dimen1}
  \caption{Claude-3.7}
\end{subfigure}}
\sbox4{\begin{subfigure}[b]{\dimen1}
  \caption{Google Translate (Chrome)}
\end{subfigure}}%
\sbox5{\begin{subfigure}[b]{\dimen1}
  \caption{ChatGPT}
\end{subfigure}}
\setlength{\dimen2}{\dimexpr \ht1+\ht2+\ht4+\dp4+\floatsep+2\lineskip}
\ifdim\ht0>\dimen2
  \def\scale{1}%
\else
  \dimen2=\ht0
  \pgfmathsetmacro{\scale}{(\ht0-\ht4-\dp4-\floatsep-2\lineskip)/(\ht1+\ht2)}%
\fi
\begin{minipage}[b]{\dimen1}
  \centering
  {\resizebox{!}{\dimen2}{\usebox0}}
  \usebox3
\end{minipage}\hfill
\begin{minipage}[b]{\dimen1}
  \centering
  {\scalebox{\scale}{\usebox1}}
  \usebox4
  \vskip\floatsep
  {\scalebox{\scale}{\usebox2}}
  \usebox5
\end{minipage}    
\caption{Examples of common online translation systems in response to gender ambiguity.}
\label{fig:translation_systems_in_practice}
\end{figure*}


Overall, the lack of alternative morphemes for unknown gender is disappointing; we observe alternative morphology ($N_5$) in only 4/21 systems (Dubformer, Claude-3.5, GPT4, and CommandR+), and usually at very low rates (less than 2\% of gender-ambiguous inputs), with the exception of Claude-3.5 in the EN$\rightarrow$ IS translation direction under active gender ambiguity. While these LLM-based\footnote{Dubformer is a proprietary system; its architecture and training details are unknown.} translation systems have the capacity and freedom to use gender-neutral morphology, we find that this is not usually invoked in practice, even for simple adjectives. Furthermore, the new morphemes that we do observe typically conform to a binary gender assumption (e.g., ``o/a'' for masculine and feminine cases), with the sole exception of Dubformer, which uses ``@'' in place of ``o/a'' in Spanish. 

Investigating further, we observe that popular solutions for casual MT users have a tendency to push binary gender solutions -- if gender ambiguity is acknowledged at all.
Referring to the examples shown in Figure \ref{fig:translation_systems_in_practice}, Google Translate tends to offer default masculine translations for lower resource languages (Icelandic), and masculine/feminine variants for higher resource languages (Spanish), but never a gender-neutral variant of the gendered adjectives. 
Both Claude-3.7 and ChatGPT provide masculine default translations given a simple translate instruction. 
However, chain-of-thought prompting can successfully be used to force the LLMs to consider the gender of the adjective's referent.  
Binary gender alternatives are produced when LLMs are instructed to consider gender, and gender-neutral strategies are used only when explicitly reminded that gender is not binary. 
When gender neutrality is requested, the specific strategies can vary. The examples show Claude-3.7 using the neuter case, and ChatGPT using an alternative part of speech (e.g., switching ``muscular'' for ``having muscles''). It does appear that prompting can be used to steer the LLMs towards a preferred gender-neutral strategy (e.g., asking ChatGPT to suggest gender-neutral synonyms for ``muscular'' produces ``fuerte (strong)''). 

Large language models in combination with prompting strategies may be a promising direction for gender-neutral machine translation, but some limitations remain.  
Although LLMs have the power to provide these nuanced explanations with multiple translation options, they often fail to give \textit{usable} translation options given the context. 
That is, providing an explanation with feminine and masculine options may be a partially \textit{correct} response, but it is neither \textit{complete} (lacking non-binary options), nor \textit{useful} given the information in the source (lacking gender neutrality given gender ambiguity). 
Additionally, the explanation provided by the LLM must be distilled to a singular output translation if used at scale. 
Ultimately, LLM-based methods should be assessed based on a single output translation based on the (limited) information provided by the source input. 

Given that machine translation systems can be steered toward a particular strategy, it becomes important to consider what the ideal gender-neutral translation should be. The availability of gender neutral adjectives, or other parts of speech, naturally varies by target language. Forcing gender neutrality can decrease translation quality, both in terms of coherence in the target language and faithfulness to the source input. Should gender neutrality in translation only be enforced in response to gender ambiguity (high gender neutral response), or should it be enforced for all inputs (high baseline gender neutrality)? A high gender neutral response indicates less overall intervention, but may introduce a disparity in translation quality for non-binary genders. 

Finally, it should be noted that the proposed metrics capture a certain aspect of gender-inclusive machine translation, but do not cover all aspects. For example, gender stereotypes can negatively affect translations while maintaining gender neutrality, and this effect is not captured by either the baseline gender neutrality or by the gender neutral response. We observe mismatched adjective translations depending on binary gender stereotypes (e.g. ``thick'' has divergent translation as either ``dumb'' (M) or ``fat'' (F), likewise ``modest''$\rightarrow$ ``humble''(M)/``covered''(F), etc.). If these divergent translations are each gender-neutral in the target language, the stereotype effect will be obscured. Therefore, gender neutrality should be considered as just one branch of ideal gender-inclusive machine translation.   


\section{Conclusion}

Gender-neutral translation is a path toward gender-inclusive machine translation that helps prevent misgendering and representational biases. However, languages with grammatical gender pose significant technical challenges. A prerequisite for MT systems is the ability to recognize when gender neutrality should be applied.
In a large-scale analysis of 21 MT systems, we evaluated their sensitivity to gender ambiguity in the source text. 
While a masculine default response is often observed, the gender-neutral response is lacking. That is, MT systems struggle to switch to gender-neutral strategies, despite the availability of such strategies in the target languages (e.g., grammatical gender neuter cases, abundant gender-neutral in-vocabulary alternatives, and an observed capacity to use alternative morphology).
Our findings highlight the need for further research in this challenging subtask of gender-inclusive translation.
Perhaps encouragingly, a small handful of systems did demonstrate a non-zero gender-neutral response, though triggered only by active gender ambiguity. 
Future work may examine the properties of these systems that enabled such a response, in order to strengthen the effect in other contexts.
Additionally, future work might examine the utility of the templated inputs used here, differing only in gender ambiguity, as a training set to improve gender-neutral translation in more complex scenarios, such as those found in the GeNTE benchmark.

\bibliography{anthology,mtsummit25}

\appendix

\section{Test Suite Details and Full Results}
\label{sec:appendix}

The full ``gender resolution in literary dialogue settings'' test suite contains various types of source inputs, each containing dialogue and meta-context surrounding the dialogue. The meta-context may include gender-stereotyped character descriptions and adverbs that control the manner of speaking. In all cases, adjectives within spoken dialogue refer to characters in the meta-context. The challenge of the test suite is to use the meta-context to correctly resolve each adjective's referent, as measured through gender agreement. 

For our purposes, we take subsets of the test suite that can be paired to create (determined, ambiguous)-gender inputs, while all other details remain constant. Test suite inputs that support the study of gender ambiguity by omission take the form: 
\begin{equation}
\label{eqn:one_person_known}
    \begin{split}
        & \text{The \{woman, man\} smiled. ``I think \{I'm, you're\} } \\
        & A_1 \text{,'' \{she, he\} said.} \\
        & \text{\{He, She\} laughed back. ``No, [\{you're, I'm\} not } \\
        & A_1 \text{, but] \{you are, I am\} } A_2 \text{,'' \{he, she\} replied.}
    \end{split}
\end{equation}

\begin{equation}
\label{eqn:two_person_known}
    \begin{split}
    & \text{The \{man, woman\} smiled. ``I think I'm } A_1 \text{ and} \\
    & \text{you're } A_2 \text{,'' \{he, she\} said.} \\
    & \text{\{He, She\} laughed back. ``No, you're } A_3 \text{, but I'm } \\
    &  A_4 \text{,'' \{he, she\} replied.} 
    \end{split}
\end{equation}

\begin{equation}
\label{eqn:one_person_partial}
    \begin{split}
        & \text{\{I, The wo/man\} smiled. ``I think \{I'm, you're\} } \\
        & A_1 \text{,'' \{I, s/he\} said.} \\
        & \text{\{S/he, I\} laughed back. ``No, [\{you're, I'm\} not } \\
        & A_1 \text{, but] \{you are, I am\} } A_2 \text{,'' \{s/he, I\} replied.}
    \end{split}
\end{equation}

\begin{equation}
\label{eqn:two_person_partial}
    \begin{split}
    & \text{\{I, The wo/man\} smiled. ``I think I'm } A_1 \text{ and} \\
    & \text{you're } A_2 \text{,'' \{I, s/he\} said.} \\
    & \text{\{S/He, I\} laughed back. ``No, you're } A_3 \text{, but I'm } \\
    &  A_4 \text{,'' \{s/he, I\} replied.} 
    \end{split}
\end{equation}
where adjectives are denoted by $A_i$, curly braces denote template variables, and square brackets indicate optional text in templates (\ref{eqn:one_person_known}) and (\ref{eqn:one_person_partial}).
Adjectives in templates (\ref{eqn:one_person_known}) and (\ref{eqn:two_person_known}) are always gender-determined ($n_1^{Det} = 2400$, $n_2^{Det} = 3840$). A perturbation in template (\ref{eqn:one_person_known}) makes the all adjectives gender-ambiguous half of the time in template (\ref{eqn:one_person_partial}) ($n_3^{Det} = 1200$, $n_3^{Amb} = 1200$), and similarly, a perturbation in template (\ref{eqn:two_person_known}) makes half of the adjectives gender-ambiguous in every instance of template (\ref{eqn:two_person_partial}) ($n_4^{Det} = 1920$, $n_4^{Amb} = 1920$).
In all cases, binary gender and the position of the first-person speaker, as applicable to the template, are balanced across the test suite. 
The full results for the proportions of gendered adjective translations in the case of determined vs.\ ambiguous gender, using these template pairs, are shown in Table \ref{tab:I_all}. The results are macro-averaged over each template type as applicable. The baseline levels of gender-neutral translations, broken down by type (Section \ref{sec:N_types}), in the gender-determined cases are shown in Table \ref{tab:N_breakdown_all}.

Test suite instances that support the study of active gender ambiguity take the form: 
\begin{equation}
\label{eqn:char_single}
\begin{split}
    & \text{The } C_g \text{ smiled. ``I think I'm } A\text{,'' } \\
    & \text{\{he, she, they\} said to the } C_{\bar{g}}. 
\end{split}
\end{equation}
where $(C_g, C_{\bar{g}})$ pairs denote binary gender stereotyped character descriptions, 
\begin{equation}
    C_g = a_g occ_g,
\end{equation}
where $a_g$ is a gender-stereotyped adjective, and $occ_g$ is a matching gender-stereotyped occupation (e.g. ``pretty nurse'' or ``strong doctor''). 
Template instances with ``he'' or ``she'' create gender-determined adjectives ($n_5^{Det} = 352$), while instances with ``they'' create gender-ambiguous adjectives ($n_5^{Amb} = 176$). 
The binary gender stereotypes are balanced across \{he, she, they\} instances. 
The full results using these template pairs are shown in Table \ref{tab:they_all}. 

Finally, the effect of binary gender stereotypes on the gender neutrality is studied using test suite instances of the form:
\begin{equation}
    \text{``I think I'm } A\text{,'' I said } [adverb].
\end{equation}
where $adverb$ is optionally used to control the manner of speaking and aligns with a socially held stereotype about binary gender (e.g., gently vs.\ brusquely). 
In total, $n_7^{neutral} = 130$ instances without any adverb are paired with $n_7^{M} = n_7^{F} = 390$ stereotyped instances, balanced by binary gender. 
\citet{dawkins-etal-2024-wmt24} find that stereotyped manner of speaking significantly impacts the choice of gender declensions between feminine and masculine forms in translation, and here we include the levels of gender-neutral translations to observe if neutrality is affected. The full results are shown in Table \ref{tab:stereo_all}.  

\begin{table*}
  \centering
  \begin{tabular}{llrrrrr}
    \hline
    \textbf{Lang.} & \textbf{System}& $(M, F, N)_{Det}$ & $(M, F, N)_{Amb}$ & $\Delta M$ & $\Delta F$ & $\Delta N$\\
    \hline
\multirow{ 14}{*}{IS}	&	AMI	&	(0.56, 0.31, 0.13)	&	(0.67, 0.19, 0.14)	&	\textbf{0.110}	&	-0.113	&	0.003 \\
	&	Aya23	&	(0.39, 0.25, 0.36)	&	(0.46, 0.21, 0.33)	&	0.066	&	-0.037	&	-0.029 \\
	&	Claude-3.5	&	(0.44, 0.42, 0.15)	&	(0.50, 0.33, 0.17)	&	0.068	&	-0.085	&	0.017 \\
	&	Dubformer	&	(0.62, 0.23, 0.16)	&	(0.69, 0.13, 0.18)	&	\textbf{0.073}	&	-0.099	&	0.026 \\
	&	GPT-4	&	(0.41, 0.41, 0.18)	&	(0.34, 0.49, 0.17)	&	-0.068	&	0.080	&	-0.012 \\
	&	IKUN	&	(0.33, 0.48, 0.19)	&	(0.24, 0.56, 0.20)	&	\textbf{-0.084}	&	0.077	&	0.007 \\
	&	IOL-Research	&	(0.46, 0.36, 0.18)	&	(0.65, 0.19, 0.16)	&	\textbf{0.195}	&	-0.180	&	-0.015 \\
	&	Llama3-70B	&	(0.47, 0.31, 0.21)	&	(0.56, 0.24, 0.20)	&	\textbf{0.086}	&	-0.074	&	-0.012 \\
	&	ONLINE-A	&	(0.62, 0.20, 0.19)	&	(0.72, 0.10, 0.18)	&	\textbf{0.100}	&	-0.092	&	-0.008 \\
	&	ONLINE-B	&	(0.55, 0.28, 0.16)	&	(0.65, 0.18, 0.17)	&	\textbf{0.103}	&	-0.108	&	0.005 \\
	&	ONLINE-G	&	(0.57, 0.15, 0.28)	&	(0.61, 0.12, 0.27)	&	0.043	&	-0.035	&	-0.008 \\
	&	TranssionMT	&	(0.55, 0.29, 0.16)	&	(0.66, 0.18, 0.17)	&	\textbf{0.104}	&	-0.110	&	0.006 \\
	&	Unbabel-Tower70B	&	(0.46, 0.39, 0.15)	&	(0.57, 0.29, 0.14)	&	\textbf{0.106}	&	-0.097	&	-0.009 \\
\hline												
\multirow{ 19}{*}{CS}	&	Aya23	&	(0.42, 0.39, 0.20)	&	(0.31, 0.51, 0.18)	&	\textbf{-0.109}	&	0.125	&	-0.016 \\
	&	CUNI-DocTransformer	&	(0.44, 0.38, 0.18)	&	(0.66, 0.17, 0.17)	&	\textbf{0.211}	&	-0.205	&	-0.006 \\
	&	CUNI-GA	&	(0.45, 0.35, 0.20)	&	(0.53, 0.29, 0.18)	&	\textbf{0.076}	&	-0.063	&	-0.013 \\
	&	CUNI-MH	&	(0.37, 0.43, 0.21)	&	(0.22, 0.60, 0.18)	&	\textbf{-0.146}	&	0.171	&	-0.025 \\
	&	Claude-3.5	&	(0.42, 0.42, 0.16)	&	(0.66, 0.19, 0.15)	&	\textbf{0.234}	&	-0.222	&	-0.012 \\
	&	CommandR+	&	(0.40, 0.42, 0.18)	&	(0.47, 0.37, 0.16)	&	0.067	&	-0.048	&	-0.019 \\
	&	GPT-4	&	(0.45, 0.42, 0.13)	&	(0.61, 0.25, 0.13)	&	\textbf{0.163}	&	-0.163	&	-0.000 \\
	&	IKUN	&	(0.43, 0.32, 0.25)	&	(0.58, 0.18, 0.24)	&	\textbf{0.149}	&	-0.139	&	-0.010 \\
	&	IOL-Research	&	(0.43, 0.42, 0.16)	&	(0.74, 0.11, 0.15)	&	\textbf{0.315}	&	-0.309	&	-0.006 \\
	&	Llama3-70B	&	(0.45, 0.35, 0.19)	&	(0.77, 0.03, 0.19)	&	\textbf{0.322}	&	-0.320	&	-0.002 \\
	&	NVIDIA-NeMo	&	(0.55, 0.28, 0.18)	&	(0.57, 0.27, 0.16)	&	0.023	&	-0.010	&	-0.013 \\
	&	ONLINE-A	&	(0.44, 0.39, 0.18)	&	(0.36, 0.49, 0.16)	&	\textbf{-0.079}	&	0.100	&	-0.021 \\
	&	ONLINE-B	&	(0.58, 0.25, 0.18)	&	(0.75, 0.07, 0.18)	&	\textbf{0.175}	&	-0.180	&	0.005 \\
	&	ONLINE-G	&	(0.54, 0.26, 0.19)	&	(0.71, 0.12, 0.18)	&	\textbf{0.163}	&	-0.146	&	-0.017 \\
	&	ONLINE-W	&	(0.44, 0.40, 0.17)	&	(0.37, 0.47, 0.16)	&	-0.065	&	0.075	&	-0.010 \\
	&	SCIR-MT	&	(0.45, 0.38, 0.18)	&	(0.47, 0.36, 0.17)	&	0.023	&	-0.014	&	-0.009 \\
	&	TranssionMT	&	(0.46, 0.36, 0.18)	&	(0.44, 0.39, 0.16)	&	-0.022	&	0.033	&	-0.011 \\
	&	Unbabel-Tower70B	&	(0.44, 0.37, 0.20)	&	(0.49, 0.33, 0.18)	&	0.056	&	-0.042	&	-0.014 \\
\hline												
\multirow{ 16}{*}{ES}	&	Aya23	&	(0.34, 0.26, 0.40)	&	(0.42, 0.22, 0.37)	&	\textbf{0.077}	&	-0.047	&	-0.030 \\
	&	Claude-3.5	&	(0.31, 0.30, 0.38)	&	(0.45, 0.18, 0.37)	&	\textbf{0.141}	&	-0.127	&	-0.014 \\
	&	CommandR+	&	(0.33, 0.31, 0.36)	&	(0.26, 0.39, 0.36)	&	\textbf{-0.072}	&	0.075	&	-0.003 \\
	&	Dubformer	&	(0.41, 0.23, 0.37)	&	(0.55, 0.09, 0.36)	&	\textbf{0.141}	&	-0.135	&	-0.006 \\
	&	GPT-4	&	(0.33, 0.29, 0.38)	&	(0.56, 0.06, 0.37)	&	\textbf{0.235}	&	-0.230	&	-0.005 \\
	&	IKUN	&	(0.29, 0.31, 0.40)	&	(0.28, 0.32, 0.40)	&	-0.013	&	0.018	&	-0.005 \\
	&	IOL-Research	&	(0.34, 0.28, 0.38)	&	(0.45, 0.17, 0.38)	&	\textbf{0.111}	&	-0.115	&	0.004 \\
	&	Llama3-70B	&	(0.35, 0.27, 0.38)	&	(0.56, 0.09, 0.36)	&	\textbf{0.208}	&	-0.186	&	-0.022 \\
	&	MSLC	&	(0.38, 0.15, 0.47)	&	(0.42, 0.12, 0.46)	&	0.044	&	-0.036	&	-0.008 \\
	&	ONLINE-A	&	(0.41, 0.21, 0.38)	&	(0.51, 0.11, 0.38)	&	\textbf{0.105}	&	-0.101	&	-0.004 \\
	&	ONLINE-B	&	(0.38, 0.22, 0.40)	&	(0.50, 0.11, 0.39)	&	\textbf{0.118}	&	-0.104	&	-0.014 \\
	&	ONLINE-G	&	(0.40, 0.20, 0.39)	&	(0.47, 0.14, 0.38)	&	0.069	&	-0.060	&	-0.009 \\
	&	ONLINE-W	&	(0.33, 0.30, 0.37)	&	(0.40, 0.23, 0.37)	&	0.066	&	-0.069	&	0.003 \\
	&	TranssionMT	&	(0.39, 0.21, 0.40)	&	(0.51, 0.11, 0.38)	&	\textbf{0.120}	&	-0.101	&	-0.019 \\
	&	Unbabel-Tower70B	&	(0.32, 0.29, 0.39)	&	(0.37, 0.23, 0.40)	&	0.046	&	-0.052	&	0.006 \\
    \hline
  \end{tabular}
  \caption{\textbf{Responses to Gender Ambiguity by Omission:} A comparison of the proportion of masculine $(M)$, feminine $(F)$, and gender-neutral translations $(N)$ when the gender of the adjective referent is known (subscript $Det$) vs.\ when it is ambiguous by omission in the source via ``I'' or ``you'' (subscript $Amb$). 
  The default masculine response is denoted by $\Delta M$, and the gender-neutral response is denoted by $\Delta N$.
  }
  \label{tab:I_all}
\end{table*}

\begin{table*}
  \centering
  \begin{tabular}{llrrrrr}
    \hline
    \textbf{Lang.} & \textbf{System}& $(M, F, N)_{Det}$ & $(M, F, N)_{Amb}$ & $\Delta M$ & $\Delta F$ & $\Delta N$\\
    \hline
\multirow{ 14}{*}{IS}	&	AMI	&	(0.46, 0.43, 0.11)	&	(0.72, 0.15, 0.13)	&	\textbf{0.262}	&	-0.282	&	0.020 \\
	&	Aya23	&	(0.50, 0.20, 0.30)	&	(0.41, 0.21, 0.38)	&	\textbf{-0.092}	&	0.014	&	\textbf{0.078} \\
	&	Claude-3.5	&	(0.42, 0.36, 0.22)	&	(0.51, 0.09, 0.40)	&	\textbf{0.087}	&	-0.267	&	\textbf{0.180} \\
	&	Dubformer	&	(0.68, 0.14, 0.19)	&	(0.72, 0.11, 0.17)	&	0.042	&	-0.026	&	-0.016 \\
	&	GPT-4	&	(0.41, 0.42, 0.18)	&	(0.51, 0.28, 0.22)	&	\textbf{0.100}	&	-0.141	&	0.041 \\
	&	IKUN	&	(0.37, 0.44, 0.19)	&	(0.41, 0.37, 0.22)	&	0.038	&	-0.07	&	0.032 \\
	&	IOL-Research	&	(0.43, 0.38, 0.19)	&	(0.68, 0.11, 0.21)	&	\textbf{0.250}	&	-0.273	&	0.023 \\
	&	Llama3-70B	&	(0.46, 0.33, 0.22)	&	(0.55, 0.21, 0.24)	&	\textbf{0.092}	&	-0.118	&	0.026 \\
	&	ONLINE-A	&	(0.46, 0.35, 0.20)	&	(0.73, 0.09, 0.19)	&	\textbf{0.270}	&	-0.258	&	-0.012 \\
	&	ONLINE-B	&	(0.45, 0.40, 0.15)	&	(0.51, 0.33, 0.16)	&	0.064	&	-0.073	&	0.009 \\
	&	ONLINE-G	&	(0.43, 0.33, 0.23)	&	(0.61, 0.16, 0.23)	&	\textbf{0.177}	&	-0.177	&	0.000 \\
	&	TranssionMT	&	(0.45, 0.40, 0.15)	&	(0.50, 0.34, 0.16)	&	0.052	&	-0.061	&	0.009 \\
	&	Unbabel-Tower70B	&	(0.44, 0.40, 0.16)	&	(0.64, 0.19, 0.16)	&	\textbf{0.199}	&	-0.206	&	0.007 \\
\hline												
\multirow{ 19}{*}{CS}	&	Aya23	&	(0.49, 0.33, 0.18)	&	(0.60, 0.23, 0.17)	&	\textbf{0.114}	&	-0.103	&	-0.011 \\
	&	CUNI-DocTransformer	&	(0.40, 0.40, 0.19)	&	(0.53, 0.22, 0.25)	&	\textbf{0.128}	&	-0.187	&	0.059 \\
	&	CUNI-GA	&	(0.40, 0.40, 0.20)	&	(0.53, 0.19, 0.28)	&	\textbf{0.128}	&	-0.209	&	\textbf{0.081} \\
	&	CUNI-MH	&	(0.36, 0.38, 0.26)	&	(0.43, 0.29, 0.29)	&	\textbf{0.071}	&	-0.094	&	0.023 \\
	&	Claude-3.5	&	(0.41, 0.43, 0.16)	&	(0.66, 0.13, 0.22)	&	\textbf{0.246}	&	-0.303	&	0.057 \\
	&	CommandR+	&	(0.39, 0.40, 0.21)	&	(0.69, 0.10, 0.22)	&	\textbf{0.297}	&	-0.3	&	0.003 \\
	&	GPT-4	&	(0.45, 0.41, 0.14)	&	(0.69, 0.14, 0.17)	&	\textbf{0.246}	&	-0.268	&	0.022 \\
	&	IKUN	&	(0.44, 0.35, 0.21)	&	(0.55, 0.22, 0.23)	&	\textbf{0.107}	&	-0.125	&	0.018 \\
	&	IOL-Research	&	(0.40, 0.40, 0.20)	&	(0.61, 0.20, 0.19)	&	\textbf{0.209}	&	-0.201	&	-0.008 \\
	&	Llama3-70B	&	(0.44, 0.36, 0.20)	&	(0.65, 0.13, 0.22)	&	\textbf{0.215}	&	-0.233	&	0.018 \\
	&	NVIDIA-NeMo	&	(0.40, 0.40, 0.19)	&	(0.62, 0.16, 0.22)	&	\textbf{0.216}	&	-0.243	&	0.027 \\
	&	ONLINE-A	&	(0.41, 0.41, 0.18)	&	(0.52, 0.30, 0.19)	&	\textbf{0.108}	&	-0.111	&	0.003 \\
	&	ONLINE-B	&	(0.41, 0.41, 0.19)	&	(0.76, 0.02, 0.22)	&	\textbf{0.349}	&	-0.384	&	0.035 \\
	&	ONLINE-G	&	(0.39, 0.39, 0.22)	&	(0.72, 0.04, 0.24)	&	\textbf{0.328}	&	-0.351	&	0.023 \\
	&	ONLINE-W	&	(0.39, 0.41, 0.19)	&	(0.48, 0.25, 0.27)	&	\textbf{0.085}	&	-0.158	&	\textbf{0.073} \\
	&	SCIR-MT	&	(0.40, 0.40, 0.20)	&	(0.56, 0.24, 0.20)	&	\textbf{0.159}	&	-0.159	&	0.000 \\
	&	TranssionMT	&	(0.40, 0.40, 0.20)	&	(0.57, 0.20, 0.22)	&	\textbf{0.170}	&	-0.198	&	0.028 \\
	&	Unbabel-Tower70B	&	(0.40, 0.40, 0.19)	&	(0.48, 0.25, 0.27)	&	\textbf{0.077}	&	-0.15	&	\textbf{0.073} \\
\hline												
\multirow{ 16}{*}{ES}	&	Aya23	&	(0.32, 0.26, 0.42)	&	(0.36, 0.22, 0.43)	&	0.039	&	-0.044	&	0.005 \\
	&	Claude-3.5	&	(0.41, 0.22, 0.37)	&	(0.51, 0.11, 0.38)	&	\textbf{0.098}	&	-0.105	&	0.007 \\
	&	CommandR+	&	(0.34, 0.29, 0.36)	&	(0.56, 0.09, 0.35)	&	\textbf{0.217}	&	-0.205	&	-0.012 \\
	&	Dubformer	&	(0.39, 0.24, 0.38)	&	(0.52, 0.11, 0.37)	&	\textbf{0.137}	&	-0.126	&	-0.011 \\
	&	GPT-4	&	(0.44, 0.18, 0.38)	&	(0.56, 0.05, 0.39)	&	\textbf{0.120}	&	-0.13	&	0.010 \\
	&	IKUN	&	(0.36, 0.21, 0.43)	&	(0.46, 0.15, 0.39)	&	\textbf{0.095}	&	-0.06	&	-0.035 \\
	&	IOL-Research	&	(0.34, 0.29, 0.38)	&	(0.49, 0.11, 0.40)	&	\textbf{0.153}	&	-0.18	&	0.027 \\
	&	Llama3-70B	&	(0.40, 0.21, 0.39)	&	(0.51, 0.13, 0.35)	&	\textbf{0.112}	&	-0.08	&	-0.032 \\
	&	MSLC	&	(0.37, 0.22, 0.41)	&	(0.53, 0.10, 0.37)	&	\textbf{0.157}	&	-0.119	&	-0.038 \\
	&	ONLINE-A	&	(0.32, 0.27, 0.40)	&	(0.56, 0.08, 0.36)	&	\textbf{0.235}	&	-0.197	&	-0.038 \\
	&	ONLINE-B	&	(0.30, 0.27, 0.43)	&	(0.54, 0.06, 0.39)	&	\textbf{0.241}	&	-0.206	&	-0.035 \\
	&	ONLINE-G	&	(0.31, 0.27, 0.42)	&	(0.51, 0.09, 0.39)	&	\textbf{0.201}	&	-0.179	&	-0.022 \\
	&	ONLINE-W	&	(0.39, 0.24, 0.38)	&	(0.51, 0.12, 0.37)	&	\textbf{0.121}	&	-0.115	&	-0.006 \\
	&	TranssionMT	&	(0.31, 0.27, 0.41)	&	(0.55, 0.08, 0.37)	&	\textbf{0.232}	&	-0.191	&	-0.041 \\
	&	Unbabel-Tower70B	&	(0.30, 0.31, 0.38)	&	(0.43, 0.17, 0.41)	&	\textbf{0.120}	&	-0.143	&	0.023 \\
    \hline
  \end{tabular}
  \caption{\textbf{Responses to Active Gender Ambiguity:} A comparison of the proportion of masculine $(M)$, feminine $(F)$, and gender-neutral translations $(N)$ when the gender of the adjective referent is known (subscript $Det$) vs.\ when it is actively ambiguous via ``they'' (subscript $Amb$). 
  The default masculine response is denoted by $\Delta M$, and the gender-neutral response is denoted by $\Delta N$.
  }
  \label{tab:they_all}
\end{table*}

\begin{table*}
  \centering
  \begin{tabular}{llrrrrrr}
    \hline
    \textbf{Lang.} & \textbf{System}& $N_{Det}$ &$N_1$ & $N_2$ & $N_3$ & $N_4$ & $N_5$  \\
    \hline
    \multirow{ 14}{*}{IS} & AMI & 0.133 & 0.125 & 0.003 & 0.003 & 0.002 & 0.000 \\
                            & Aya23 & 0.360 & 0.186 & 0.122 & 0.030 & 0.022 & 0.000 \\
                            & Claude-3.5 & 0.149 & 0.132 & 0.009 & 0.004 & 0.000 & 0.003 \\
                            & Dubformer & 0.158 & 0.111 & 0.037 & 0.002 & 0.003 & 0.005 \\
                            & GPT-4 & 0.177 & 0.144 & 0.030 & 0.002 & 0.001 & 0.000 \\
                            & IKUN & 0.191 & 0.178 & 0.008 & 0.002 & 0.002 & 0.000 \\
                            & IOL-Research & 0.177 & 0.147 & 0.012 & 0.001 & 0.016 & 0.000 \\
                            & Llama3-70B & 0.213 & 0.147 & 0.046 & 0.011 & 0.009 & 0.000 \\
                            & ONLINE-A & 0.185 & 0.141 & 0.008 & 0.004 & 0.033 & 0.000 \\
                            & ONLINE-B & 0.163 & 0.155 & 0.004 & 0.004 & 0.000 & 0.000 \\
                            & ONLINE-G & 0.279 & 0.153 & 0.039 & 0.001 & 0.086 & 0.000 \\
                            & TranssionMT & 0.163 & 0.154 & 0.004 & 0.005 & 0.000 & 0.000 \\
                            & Unbabel-Tower70B & 0.151 & 0.139 & 0.007 & 0.004 & 0.001 & 0.000 \\
                        
     \hline
    \multirow{ 19}{*}{CS} & Aya23 & 0.196 & 0.132 & 0.001 & 0.044 & 0.019 & 0.000 \\
                            & CUNI-DocTransformer & 0.180 & 0.089 & 0.001 & 0.065 & 0.025 & 0.000 \\
                            & CUNI-GA & 0.197 & 0.088 & 0.001 & 0.089 & 0.019 & 0.000 \\
                            & CUNI-MH & 0.206 & 0.099 & 0.000 & 0.087 & 0.019 & 0.000 \\
                            & Claude-3.5 & 0.161 & 0.095 & 0.000 & 0.048 & 0.018 & 0.000 \\
                            & CommandR+ & 0.180 & 0.090 & 0.000 & 0.068 & 0.022 & 0.000 \\
                            & GPT-4 & 0.132 & 0.084 & 0.000 & 0.031 & 0.018 & 0.000 \\
                            & IKUN & 0.246 & 0.129 & 0.003 & 0.089 & 0.026 & 0.000 \\
                            & IOL-Research & 0.155 & 0.082 & 0.000 & 0.057 & 0.017 & 0.000 \\
                            & Llama3-70B & 0.194 & 0.145 & 0.003 & 0.020 & 0.026 & 0.000 \\
                            & NVIDIA-NeMo & 0.175 & 0.108 & 0.002 & 0.047 & 0.019 & 0.000 \\
                            & ONLINE-A & 0.177 & 0.094 & 0.002 & 0.060 & 0.021 & 0.000 \\
                            & ONLINE-B & 0.179 & 0.098 & 0.001 & 0.060 & 0.021 & 0.000 \\
                            & ONLINE-G & 0.193 & 0.127 & 0.000 & 0.039 & 0.027 & 0.000 \\
                            & ONLINE-W & 0.169 & 0.088 & 0.000 & 0.063 & 0.018 & 0.000 \\
                            & SCIR-MT & 0.177 & 0.109 & 0.000 & 0.046 & 0.022 & 0.000 \\
                            & TranssionMT & 0.175 & 0.100 & 0.002 & 0.054 & 0.019 & 0.000 \\
                            & Unbabel-Tower70B & 0.196 & 0.106 & 0.000 & 0.073 & 0.017 & 0.000 \\
                        \hline
  \multirow{ 16}{*}{ES} & Aya23 & 0.396 & 0.367 & 0.000 & 0.021 & 0.009 & 0.000 \\
                        & Claude-3.5 & 0.383 & 0.372 & 0.000 & 0.011 & 0.000 & 0.000 \\
                        & CommandR+ & 0.359 & 0.341 & 0.000 & 0.016 & 0.002 & 0.000 \\
                        & Dubformer & 0.367 & 0.345 & 0.000 & 0.012 & 0.005 & 0.005 \\
                        & GPT-4 & 0.379 & 0.362 & 0.000 & 0.015 & 0.002 & 0.000 \\
                        & IKUN & 0.401 & 0.375 & 0.000 & 0.021 & 0.004 & 0.000 \\
                        & IOL-Research & 0.379 & 0.360 & 0.000 & 0.018 & 0.001 & 0.000 \\
                        & Llama3-70B & 0.378 & 0.353 & 0.000 & 0.020 & 0.005 & 0.000 \\
                        & MSLC & 0.471 & 0.392 & 0.000 & 0.044 & 0.035 & 0.000 \\
                        & ONLINE-A & 0.382 & 0.340 & 0.000 & 0.036 & 0.005 & 0.000 \\
                        & ONLINE-B & 0.403 & 0.351 & 0.000 & 0.047 & 0.006 & 0.000 \\
                        & ONLINE-G & 0.393 & 0.345 & 0.000 & 0.043 & 0.006 & 0.000 \\
                        & ONLINE-W & 0.367 & 0.321 & 0.000 & 0.039 & 0.007 & 0.000 \\
                        & TranssionMT & 0.399 & 0.350 & 0.000 & 0.043 & 0.006 & 0.000 \\
                        & Unbabel-Tower70B & 0.389 & 0.360 & 0.000 & 0.029 & 0.001 & 0.000 \\
    \hline
  \end{tabular}
  \caption{The \textbf{baseline gender neutrality} $(N_{Det})$ by type $(N_{\{1...5\}})$ for all translation systems for the three target languages. }
  \label{tab:N_breakdown_all}
\end{table*}

\begin{table*}
  \centering
  \small
  \begin{tabular}{llrrrrr}
    \hline
     & \textbf{System}& $(M, F, N)_{Neutral}$ & $(M, F, N)_{StereoM}$ & $(M, F, N)_{StereoF}$ & $\Delta G_{avg}$ & $\Delta N_{avg}$\\
    \hline
    \multirow{ 14}{*}{IS} & AMI & (0.69, 0.15, 0.16) & (0.75, 0.09, 0.16) & (0.74, 0.09, 0.17) & 0.001 & 0.005 \\
                        & Aya23 & (0.44, 0.26, 0.30) & (0.45, 0.20, 0.34) & (0.38, 0.27, 0.35) & 0.010 & 0.048 \\
                        & Claude-3.5 & (0.56, 0.12, 0.32) & (0.70, 0.05, 0.25) & (0.52, 0.17, 0.31) & \textbf{0.093} & -0.037 \\
                        & Dubformer & (0.70, 0.10, 0.20) & (0.65, 0.14, 0.20) & (0.62, 0.15, 0.23) & 0.003 & 0.017 \\
                        & GPT-4 & (0.29, 0.48, 0.23) & (0.47, 0.32, 0.21) & (0.24, 0.54, 0.22) & \textbf{0.119} & -0.014 \\
                        & IKUN & (0.15, 0.56, 0.28) & (0.24, 0.48, 0.28) & (0.15, 0.58, 0.27) & 0.051 & -0.010 \\
                        & IOL-Research & (0.56, 0.17, 0.27) & (0.60, 0.12, 0.27) & (0.53, 0.19, 0.27) & 0.033 & 0.005 \\
                        & Llama3-70B & (0.57, 0.17, 0.27) & (0.65, 0.11, 0.24) & (0.50, 0.23, 0.28) & \textbf{0.070} & -0.008 \\
                        & ONLINE-A & (0.63, 0.09, 0.28) & (0.62, 0.09, 0.29) & (0.62, 0.09, 0.30) & -0.008 & 0.017 \\
                        & ONLINE-B & (0.55, 0.20, 0.25) & (0.58, 0.19, 0.23) & (0.58, 0.19, 0.22) & 0.013 & -0.028 \\
                        & ONLINE-G & (0.44, 0.15, 0.41) & (0.45, 0.14, 0.41) & (0.47, 0.16, 0.37) & 0.006 & -0.017 \\
                        & TranssionMT & (0.55, 0.20, 0.25) & (0.58, 0.20, 0.22) & (0.58, 0.20, 0.22) & 0.019 & -0.036 \\
                        & Unbabel-Tower & (0.54, 0.22, 0.24) & (0.61, 0.16, 0.23) & (0.42, 0.33, 0.25) & \textbf{0.087} & 0.003 \\                    
                        \hline
    \multirow{ 19}{*}{CS} & Aya23 & (0.49, 0.33, 0.18) & (0.64, 0.17, 0.19) & (0.33, 0.50, 0.18) & \textbf{0.160} & 0.005 \\
                        & CUNI-Doc & (0.51, 0.29, 0.20) & (0.54, 0.27, 0.19) & (0.39, 0.43, 0.19) & \textbf{0.084} & -0.014 \\
                        & CUNI-GA & (0.48, 0.34, 0.18) & (0.49, 0.34, 0.17) & (0.42, 0.43, 0.15) & 0.053 & -0.020 \\
                        & CUNI-MH & (0.05, 0.76, 0.19) & (0.44, 0.32, 0.24) & (0.05, 0.75, 0.21) & \textbf{0.189} & 0.031 \\
                        & Claude-3.5 & (0.64, 0.18, 0.17) & (0.78, 0.06, 0.16) & (0.48, 0.35, 0.17) & \textbf{0.147} & -0.005 \\
                        & CommandR+ & (0.47, 0.36, 0.17) & (0.66, 0.12, 0.22) & (0.34, 0.48, 0.18) & \textbf{0.159} & 0.024 \\
                        & GPT-4 & (0.82, 0.05, 0.13) & (0.85, 0.02, 0.14) & (0.73, 0.15, 0.13) & 0.061 & 0.001 \\
                        & IKUN & (0.53, 0.27, 0.20) & (0.64, 0.15, 0.22) & (0.54, 0.27, 0.19) & 0.053 & 0.003 \\
                        & IOL-Research & (0.74, 0.11, 0.15) & (0.78, 0.07, 0.15) & (0.69, 0.17, 0.15) & 0.048 & -0.001 \\
                        & Llama3-70B & (0.73, 0.04, 0.23) & (0.75, 0.04, 0.22) & (0.70, 0.10, 0.21) & 0.037 & -0.019 \\
                        & NVIDIA-NeMo & (0.57, 0.22, 0.21) & (0.61, 0.23, 0.16) & (0.59, 0.25, 0.16) & 0.035 & -0.048 \\
                        & ONLINE-A & (0.14, 0.72, 0.14) & (0.01, 0.85, 0.13) & (0.01, 0.87, 0.12) & 0.011 & -0.012 \\
                        & ONLINE-B & (0.82, 0.02, 0.16) & (0.78, 0.06, 0.16) & (0.76, 0.07, 0.17) & 0.007 & 0.003 \\
                        & ONLINE-G & (0.77, 0.07, 0.17) & (0.76, 0.07, 0.17) & (0.78, 0.05, 0.17) & -0.010 & 0.004 \\
                        & ONLINE-W & (0.48, 0.37, 0.15) & (0.47, 0.37, 0.16) & (0.12, 0.72, 0.17) & \textbf{0.170} & 0.009 \\
                        & SCIR-MT & (0.61, 0.18, 0.21) & (0.65, 0.14, 0.21) & (0.50, 0.31, 0.19) & \textbf{0.081} & -0.007 \\
                        & TranssionMT & (0.17, 0.69, 0.14) & (0.20, 0.67, 0.13) & (0.08, 0.80, 0.12) & \textbf{0.074} & -0.016 \\
                        & Unbabel-Tower & (0.45, 0.35, 0.20) & (0.58, 0.19, 0.22) & (0.38, 0.42, 0.20) & \textbf{0.100} & 0.017 \\
                        \hline
  \multirow{ 16}{*}{ES} & Aya23 & (0.46, 0.24, 0.30) & (0.56, 0.11, 0.33) & (0.35, 0.35, 0.30) & \textbf{0.107} & 0.010 \\
                        & Claude-3.5 & (0.56, 0.12, 0.32) & (0.64, 0.03, 0.33) & (0.52, 0.16, 0.32) & 0.055 & 0.010 \\
                        & CommandR+ & (0.40, 0.30, 0.30) & (0.53, 0.17, 0.30) & (0.31, 0.40, 0.29) & \textbf{0.112} & -0.005 \\
                        & Dubformer & (0.58, 0.13, 0.28) & (0.62, 0.08, 0.30) & (0.58, 0.11, 0.31) & 0.006 & 0.018 \\
                        & GPT-4 & (0.68, 0.02, 0.29) & (0.68, 0.02, 0.30) & (0.66, 0.05, 0.29) & 0.013 & 0.003 \\
                        & IKUN & (0.27, 0.41, 0.32) & (0.34, 0.35, 0.31) & (0.26, 0.41, 0.33) & 0.036 & 0.003 \\
                        & IOL-Research & (0.59, 0.10, 0.31) & (0.65, 0.05, 0.30) & (0.58, 0.14, 0.29) & 0.049 & -0.018 \\
                        & Llama3-70B & (0.64, 0.04, 0.33) & (0.69, 0.01, 0.30) & (0.59, 0.11, 0.30) & 0.064 & -0.024 \\
                        & MSLC & (0.53, 0.09, 0.39) & (0.55, 0.08, 0.38) & (0.53, 0.10, 0.38) & 0.018 & -0.012 \\
                        & ONLINE-A & (0.55, 0.09, 0.35) & (0.55, 0.09, 0.36) & (0.52, 0.12, 0.36) & 0.009 & 0.007 \\
                        & ONLINE-B & (0.59, 0.08, 0.34) & (0.59, 0.07, 0.34) & (0.57, 0.09, 0.34) & 0.008 & 0.001 \\
                        & ONLINE-G & (0.51, 0.15, 0.34) & (0.50, 0.16, 0.35) & (0.50, 0.15, 0.35) & -0.003 & 0.006 \\
                        & ONLINE-W & (0.48, 0.16, 0.36) & (0.51, 0.12, 0.36) & (0.28, 0.36, 0.36) & \textbf{0.116} & 0.001 \\
                        & TranssionMT & (0.56, 0.10, 0.34) & (0.55, 0.10, 0.35) & (0.52, 0.13, 0.34) & 0.011 & 0.010 \\
                        & Unbabel-Tower & (0.42, 0.22, 0.36) & (0.47, 0.16, 0.37) & (0.34, 0.32, 0.33) & \textbf{0.077} & -0.011 \\
    \hline
  \end{tabular}
  \caption{Effect of binary \textbf{gender stereotypes} on the proportion of masculine $(M)$, feminine $(F)$, and gender-neutral translations $(N)$. Subscripts denote the stereotype influencing the assumed gender of the adjective referent. The average effect on binary gender is denoted by $\Delta G_{avg}$, and the average effect on the proportion of neutral translations is denoted by $\Delta N_{avg}$.}
  \label{tab:stereo_all}
\end{table*}

\end{document}